\def\year{2020}\relax
\documentclass[letterpaper]{article} 
\usepackage{aaai20}  
\usepackage{times}  
\usepackage{helvet} 
\usepackage{courier}  
\usepackage[hyphens]{url}  
\usepackage{graphicx} 
\usepackage{graphicx}  
\usepackage{amsmath}
\usepackage{amsfonts}
\usepackage{booktabs}
\usepackage{multirow}

\title{Feedback-Based Self-Learning in Large-Scale Conversational AI Agents}
\author{
    Pragaash Ponnusamy, Alireza Roshan Ghias, Chenlei Guo, Ruhi Sarikaya\\ 
    Amazon Alexa\\
    ponnup@amazon.com, ghiasali@amazon.com, guochenl@amazon.com, rsarikay@amazon.com
}
 
\begin{document}

\maketitle

\begin{abstract}
    
    Today, most of the large-scale conversational AI agents such as Alexa, Siri, or Google Assistant are built using manually annotated data to train the different components of the system including Automatic Speech Recognition (ASR), Natural Language Understanding (NLU) and Entity Resolution (ER). Typically, the accuracy of the machine learning models in these components are improved by manually transcribing and annotating data. As the scope of these systems increase to cover more scenarios and domains, manual annotation to improve the accuracy of these components becomes prohibitively costly and time consuming.
    In this paper, we propose a system that leverages customer/system interaction feedback signals to automate learning without any manual annotation. Users of these systems tend to modify a previous query in hopes of fixing an error in the previous turn to get the right results. These reformulations, which are often preceded by defective experiences caused by either errors in ASR, NLU, ER or the application. In some cases, users may not properly formulate their requests (e.g. providing partial title of a song), but gleaning across a wider pool of users and sessions reveals the underlying recurrent patterns. Our proposed self-learning system automatically detects the errors, generate reformulations and deploys fixes to the runtime system to correct different types of errors occurring in different components of the system. In particular, we propose leveraging an absorbing Markov Chain model as a collaborative filtering mechanism in a novel attempt to mine these patterns. We show that our approach is highly scalable, and able to learn reformulations that reduce Alexa-user errors by pooling anonymized data across millions of customers. The proposed self-learning system achieves a win/loss ratio of 11.8 and effectively reduces the defect rate by more than 30\% on utterance level reformulations in our production A/B tests. To the best of our knowledge, this is the first self-learning large-scale conversational AI system in production.
    
\end{abstract}

\section{Introduction}
Large-scale conversational AI agents such as Alexa, Siri, and Google Assistant are getting more and more prevalent, opening up in new domains and taking up new tasks to help users across the globe. One key consideration in designing such systems is how they can be improved over time at that scale. Users interacting with these agents experience frictions due to various reasons: 1) Automatic Speech Recognition (ASR) errors, such as \textit{"play maj and dragons"} (should be \textit{"play imagine dragons"}), 2) Natural Language Understanding (NLU) errors, such as \textit{"don't play this song again skip"} (Alexa would understand if it is formulated as \textit{"thumbs down this song"}), and even user errors, such as \textit{"play bazzi angel"} (it should've been \textit{"play beautiful by bazzi"}). It goes without saying that fixing these frictions help users to have a more seamless experience, and engage more with the AI agents.

\begin{figure*}[t]
    \centering
    \includegraphics[width=0.8\textwidth]{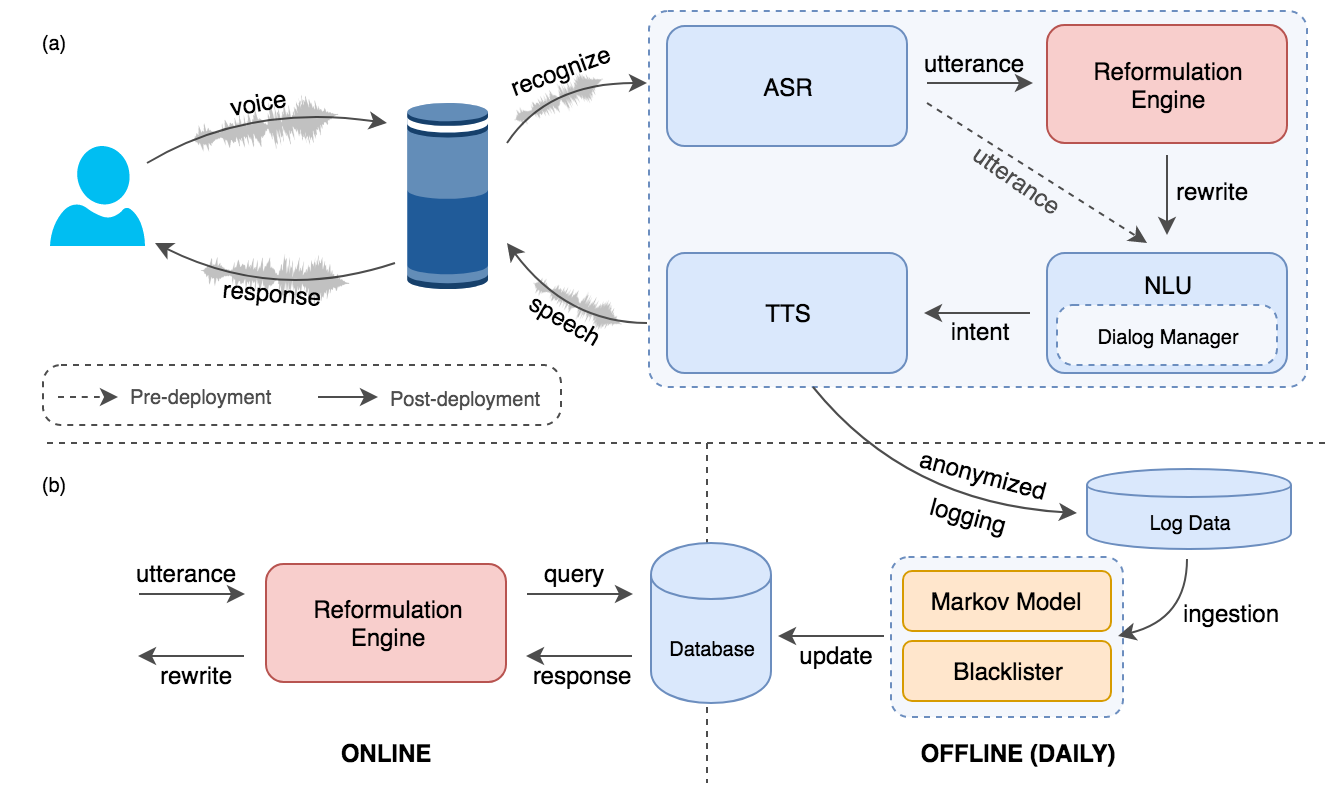}
    \caption{A high-level overview of the deployed architecture with our reformulation engine in context of the overall system in (a) and the offline sub-system that updates its online counterpart on a daily cadence.}
    \label{fig:archfig}
\end{figure*}

One common method to address frictions is to gather these use cases and fix them manually using rules and Finite State Transducers (FST) as they're often the case in speech recognition systems \cite{mohri}. This of course is a laborious technique which is: 1) not scalable at Alexa scale, and 2) prone to error, and 3) getting stale and even defective over time. Another approach could be to identify these frictions, ask annotators to come up with the correct form of query, and then update ASR and NLU models to solve these problems. This is also: 1) not an scalable solution, since it needs a lot of annotations, and 2) it is expensive and time consuming to update those models. Instead, we have taken a "query rewriting" approach to solve customer frictions, meaning that when necessary, we reformulate a customer's query such that it conveys the same meaning/intent, and is actionable (i.e. interpretable) by Alexa's existing NLU systems.

In motivating our approach, consider the example utterance, \textit{"play maj and dragons"}. Now, without reformulation, Alexa would inevitably come up with the response, \textit{"Sorry, I couldn't find maj and dragons"}. Some customers give up at this point, while others may try enunciating better for Alexa to understand them: \textit{"play imagine dragons"}. Also note that there might be other customers who give up, and change the next query to another intent, for example: \textit{"play pop music"}. Here, frictions evidently cause dissatisfaction with different customers reacting differently to them. However, quite clearly there are good rephrases by some customers among all these interactions, which beckons the question -- how can we identify and extract them to solve customer frictions?

We propose using a Markov-based collaborative filtering approach to identify rewrites that lead to successful customer interactions. We go on to discuss the theory and implementation of the idea, as well as show that this method is highly scalable and effective in significantly reducing customer frictions. We also discuss how this approach was deployed into customer-facing production and what are some of the challenges and benefits of such approach.

\section{Related Work}
Collaborative filtering has been used extensively in recommender systems. In a more general sense, collaborative filtering can be viewed as a method of mining patterns from various agents (most commonly, people), in order to help them help each other out \cite{terveen}. Markov chains have been used previously in collaborative filtering applications to recommend course enrollment \cite{khorasani}, personalized recommender systems \cite{sahoo}, and web recommendation \cite{fouss}. 

Studies have shown that Markov processes can be used to explain the user web query behavior \cite{jansen}, and Markov chains have since been used successfully for web query reformulation via absorbing random walk \cite{wang}, and modeling query utility \cite{zhu}. We here present a new method for query reformulation using Markov chain that is both highly scalable and interpretable due to intuitive definitions of transition probabilities. Also, to the best of the authors' knowledge, this is the first work where Markov chain is used for query reformulation in voice-based virtual assistants.

One important difference between the web query reformulation and Alexa's use case is that we need to seamlessly replace the user's utterance in order to remove friction. Asking users for confirmation every time we plan to reformulate is on itself an added friction, which we try to avoid as much as possible. Another difference is how success and failure are defined for an interaction between user and a voice-based virtual assistant system. We use implicit and explicit user feedback when interacting with Alexa to establish the \textit{absorbing states} of success and failure.

\section{System Overview}
The Alexa conversational AI system follows a rather well-established architectural pattern of cloud-based digital voice assistants \cite{jianfeng} i.e. comprising of an automatic speech recognition (ASR) system, a natural language understanding (NLU) system with a built-in dialog manager, and a text-to-speech (TTS) system, as visualized in Fig. \ref{fig:archfig}. Conventionally, as a user interacts with their Alexa-enabled device, their voice is first recognized by ASR and decoded into plain text, which we refer to as an utterance. The utterance is then interpreted by the NLU component to surface the aforementioned user's intent by also accounting for the state of user's active dialog session. Thereafter, the intent and the corresponding action to execute is passed on to the TTS to generate the appropriate response as speech back to the user via their Alexa-enabled device, thus closing the interaction loop. Also note that the metadata associated with each of the above systems are anonymized and logged asynchronously to an external database.

In deploying our self-learning system, we first intercept the utterance being passed onto the NLU system and rewrite it with our reformulation engine. We then subsequently pass the rewrite in lieu of the original utterance back to NLU for interpretation, and thus restoring the original data flow. This is shown as the post-deployment data flow path in Fig. \ref{fig:archfig}. Our reformulation engine is essentially implements rather lightweight service-oriented architecture that encapsulates the access to a high-performance, low-latency database, which is queried with the original utterance to yield its corresponding rewrite candidate. This along with the fact that the system is fundamentally stateless across users translates to a rather scalable customer-facing system with marginal impact to the user perceived latency of their Alexa-enabled device. 

In order to discover new rewrite candidates and maintain the viability of existing rewrites, our Markov-based model ingests the anonymized Alexa log data on a daily basis to learn from users' reformulations and subsequently updates the aforementioned online database. We discuss the nature of the dataset and how our model achieves this in later sections of this paper. This ingestion to update process takes place offline in entirety with the rewrites in the database updated via a low-maintenance feature-toggling (i.e. feature-flag) mechanism. Additionally, we also have an offline blacklisting mechanism which evaluates the rewrites from our Markov model by independently comparing their friction rate against that of the original utterance, and subsequently filtering them from being uploaded to the database should they perform worse against their no-rewrite counterpart using a $Z$-test with a rather conservative $p$-value of $0.01$. This allows us to maintain a high precision system at runtime. It is worth mentioning that friction detection is done using a pre-trained ML model based on user's utterance and Alexa's response. The details of that model is out of scope of this paper.

\section{Dataset}
As our objective is to learn the patterns from user interactions with Alexa, we pre-process 3 months of anonymized Alexa log data across millions of customers, which constitutes a highly randomized collection of time-series utterance data, to build our dataset, $\mathcal{D}$ comprising of a set of sessions, $S$ i.e.: 

\begin{equation}
    \mathcal{D} = \lbrace S_0, S_1, \ldots \rbrace
\end{equation}

Here, in defining the concept of a session, we first define the construction function $f$, parameterized by a customer, $c$, a device, $d$, and an initial timestamp, $\tau_0$, to yield a finite ordered set of successive utterances, $u$ (\textit{and its associated metadata}) such that the time delay between any two consecutive utterances is at most $\delta_{\tau}$. We also note that interjecting utterances, $J$, i.e. those leading to \texttt{StopIntent}, \texttt{CancelIntent}, etc., that occur before the end of the aforementioned set are removed. Then, a session, $S_k$ is defined as follows:

\begin{equation}
    S_k =  f(c, d, \tau_0) = \left( u_0^{(k)}, u_1^{(k)}, \ldots, u_{T_k}^{(k)} \right)
\end{equation}

\noindent such that the following properties hold true:

\begin{itemize}
    \item $\tau_0^{(k)} = \tau_0$, and 
    \item $\tau_j^{(k)} > \tau_i^{(k)},\ \forall\ 0 \leq i < j \leq T_k$, and
    \item $\Bigr| \tau_{i+1}^{(k)} - \tau_i^{(k)} \Bigr| \leq \delta_{\tau}$, and 
    \item $u_i \not\in J,\ \forall\ i < T_k$.
\end{itemize}

\begin{figure*}[t]
    \centering
    \includegraphics[width=0.9\textwidth]{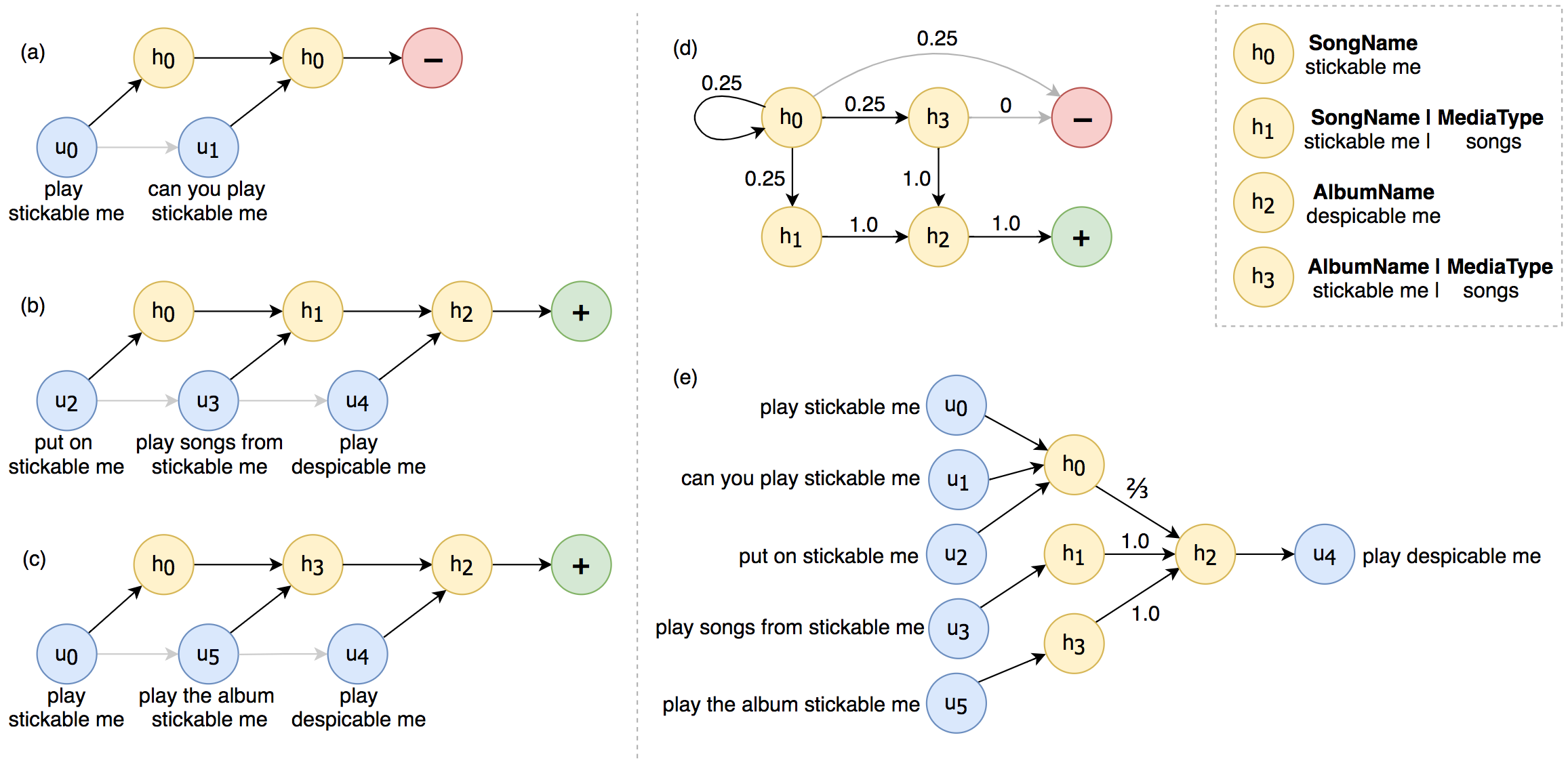}
    \caption{A visual representation of the Markov model constructed in the interpretation space, $H$, over three separate sessions, (a), (b), and (c), of users attempting to play the album \textit{"Despicable Me"}, and how solving for the path with the highest likelihood of success, (+), given by the darkened edged in (d), can allow for the defective utterances to be reformulated into a more successful query, as summarized in (e). Note that here, for demonstration purposes, we only show 3 interactions. However, in practice, we had a higher threshold for the minimum number of customers and interactions to have better estimates for the probabilities.}
    \label{fig:graphfig}
\end{figure*}

Intuitively speaking, a session is effectively a time-delimited snapshot of a user's conversation history with their Alexa-enabled device. We illustrate this in Fig. \ref{fig:graphfig} (a), (b), and (c) where each session is represented as a linear directed chain of successive utterances e.g. $u_2 \rightarrow u_3  \rightarrow u_4$. In this paper, we choose the value of $\delta_{\tau} = 45$ seconds as a result from a separate study.

\section{Absorbing Markov Chain}

In this section, we show how encoding user interaction history as paths in an absorbing Markov Chain model can be used to mine patterns for reformulating utterances. In particular, we discuss in detail the concept of the interpretation space, $H$ (Section 4.1), which functions as the vertex set of the model's transient states (Section 4.2). We then elaborate on the construction of the absorbing states, $R$ (Section 4.3), the canonical solution to the model (Section 4.4), and the practical implementation of the model (Section 4.5). As the Markov Chain model is inherently a probabilistic graphical model, we can represent it as graph, $G = (V, E)$, where the vertex set, $V$ and the edge set, $E$ are given as follows:

\begin{equation}
    V = H \cup R \hspace{0.3in} E = \left\lbrace (x, y)\ |\ x \in H \land y \in V \right\rbrace
\end{equation}

We note that from here on out, we use the terms, Graph and Markov model interchangeably.

\subsection{Interpretation Space}
While our definition of a session in Section 3 naturally extends towards having each ordered linear sequence of utterances as a path in our Markov model, this encoding in the utterance space, $U$ i.e. the space of all utterances $u$, imposes a limitation on the model by creating heavily sparse connections. This is primarily due to the high degree of semantic and structural variance in $U$, which would ultimately result in a lower capacity for generalization.

To resolve this, we leverage the domain and intent classifier as well as the named entity recognition (NER) results from Alexa's NLU systems to surface structured representations of utterances, and thus encapsulate a latent distribution over $U$. Consequently, each utterance in a session is projected into this interpretation space, $H$ which comprises the set of all interpretations $h$, to define a latent session:

\begin{equation}
    S_k^{\prime} = \left( h_0^{(k)}, h_1^{(k)}, \ldots, h_{T_k}^{(k)} \right)
\end{equation}

To exemplify this, consider the utterance, \textit{"play \textbf{despicable me}"} (i.e. $u_4$ in Fig. \ref{fig:graphfig}), which would be mapped into the $H$-space as:

\begin{center}
    $\textsf{Music} \vert \textsf{PlayMusicIntent} \vert \textsf{AlbumName:}\textbf{\textit{\textsf{despicable me}}}$
\end{center}

\noindent which is compactly represented as $h_2$ in Fig. \ref{fig:graphfig}. As the $H$-space condenses the semantics of $U$, this mapping between $U$ and $H$ is inherently a many-to-one relationship. However, given the stochasticity of Alexa's NLU, the original projection itself is not entirely bijective and thus results in a many-to-one relationship in both the forward and inverse mapping, i.e. $U \rightarrow H$ and $H \rightarrow U$, akin to a bipartite mapping. This in turn, yields the conditional probability distributions, $P(H | U)$ and $P(U | H)$, such that for a particular $u \in U$ and $h \in H$, they are defined as follows:

\begin{equation}
    P\left(h | u \right) = \frac{c\left(u, h\right)}{\sum\limits_{h^{\prime} \in H}c\left(u, h^{\prime}\right)} \hspace{0.3in} P\left(u | h \right) = \frac{c\left(u, h\right)}{\sum\limits_{u^{\prime} \in U}c\left(u^{\prime}, h\right)}
\end{equation}

\noindent where $c(u, h)$ is the co-occurrence count of the pair $\left(u, h\right)$ in the dataset, $\mathcal{D}$ i.e. the total number of times both $u$ and $h$ are mapped onto each other.

\subsection{Transient States}
Given our transformed dataset, $\mathcal{D}^{\prime}$ of latent sessions $S^{\prime}$, we take each such session and the interpretations within it to represent paths and transient states respectively in our Markov model, such that each successive pair of interpretations would represent an edge in the Graph. In defining the transition probability distribution, we first define $Z_i$, the total occurrence of an interpretation $h_i$ in the aforementioned dataset as follows:

\begin{equation}
    Z_i = \sum\limits_{v \in V} c\left(h_i, v\right)
\end{equation}

\noindent where $c\left(h_i, v\right)$ is the co-occurrence count of the pair $\left(h_i, v\right)$ i.e. the total number of times $v$ is adjacent to $h_i$, aggregated across all sessions (\textit{i.e. over 3 months and millions of customers}) in $\mathcal{D^{\prime}}$:

\begin{equation}
    c\left( h_i, v \right) = \sum\limits_k \sum\limits_{t=0}^{T_k - 1} \mathbf{1}\left( h_t^{(k)} = h_i \land h_{t+1}^{(k)} = v \right)
\end{equation}

\noindent Then, the corresponding probability that a transition state $h_i \in H$ transitions to $h_j \in H$ in the Graph is given by:

\begin{equation}
    P\left( h_j | h_i \right) = \frac{c\left( h_i, h_j \right)}{Z_i}
\end{equation}

Taking this in context of Fig. \ref{fig:graphfig}, consider the transition probability $P\left( h_1 | h_0 \right)$. From the sessions (a), (b), and (c), we can note that the transition state $h_0$ is adjacent to the states, $\lbrace h_0, h_1, h_3, (-) \rbrace$ with each of them having a co-occurrence of $1$ with $h_0$. Here, $(-)$ refers to the failure absorbing state (\textit{defined in the following sub-section}). As such, the probability $P\left( h_1 | h_0 \right) = \frac{1}{4} = 0.25$ as shown in (d).

\subsection{Absorbing States}
In formulating the definition of the absorbing states of the Markov model, we look towards encoding the notion of interpreted defects as perceived by the user. As we have briefly introduced earlier, this concept of defect surfaces in two key forms i.e. via \textit{explicit} and \textit{implicit} feedback.

Here, \textit{explicit feedback} refers to the type of corrective or reinforcing feedback received from direct user engagement. This primarily includes events where users opt to interrupt Alexa by means of an interjecting utterance (\textit{as defined above in Section 3}). This is illustrated in the example below:

\begin{center}
    \begin{tabular}{rl}
        \textbf{User:} & \textit{"play a lever"}  \\
        \textbf{Alexa:} & \textit{"Here's Lever by The Mavis's, starting now."} \\
        \textbf{User:} & \textit{"stop"}
    \end{tabular}
\end{center}

In contrast, \textit{implicit feedback} is typically observed when users abandon a session following Alexa's failure to handle a request either due to an internal exception or simply unable to find a match for the entities resolved. Case in point: 

\begin{center}
    \begin{tabular}{rl}
        \textbf{User:} & \textit{"play maj and dragons"}  \\
        \textbf{Alexa:} & \textit{"Sorry, I can't find the artist maj and dragons."} 
    \end{tabular}
\end{center}

Given this, we define two absorbing states: \textit{failure} ($r^-$), and \textit{success} ($r^+$), where \textit{success} is defined as the absence of \textit{failure}. These states are artificially injected to the end of \textit{all} sessions, based on the implicit and explicit feedback we infer from Alexa's response, and user's last utterance. 

To clarify this, let's walk through the examples above assuming that they are the last utterances of their corresponding sessions. In the first example, we would drop the \textit{"stop"} turn, and add a \textit{failure} state. In the second example, we simply add the \textit{failure} state to the end of the session. Finally, in the absence of an explicit or implicit feedback, we add a \textit{success} state to the end of the session. There are certain edge cases, but for the sake of brevity, we do not discuss them here. Given this, we can then define the probability that a given transient state, $h_i$ is absorbed in much the same way as in Eq. 8, e.g.:

\begin{equation}
    P\left( r^+ | h_i \right) = \frac{c\left( h_i, r^+ \right)}{Z_i}
\end{equation}

Note that in Fig. \ref{fig:graphfig}, we refer to the \textit{failure} ($r^-$), and \textit{success} ($r^+$) states as $(-)$ and $(+)$ respectively.

\subsection{Markov Model}
With the distributions over both the transition and absorbing states defined above, recall that the interpretation space, $H$ is the set of all transient states in the Graph. Then, we can summarize the Markov model in its canonical form via the transition matrix, $\mathbf{A}$ as follows:

\begin{equation}
    \mathbf{A} = \begin{bmatrix}
        \mathbf{Q} & \mathbf{R} \\
        \mathbf{0} & \mathbf{I}_2
    \end{bmatrix}
\end{equation}

\noindent where:
\begin{itemize}
    \item $\mathbf{Q} \in \mathbb{R}^{|H| \times |H|}$ s.t. $q_{i,j} = P\left(h_j | h_i \right)$,
    \item $\mathbf{R} = \left[ \mathbf{r}^+, \mathbf{r}^-\right] \in \mathbb{R}^{|H| \times 2}$ s.t. $r_i = \left[ P\left( r^+ | h_i \right), P\left( r^- | h_i \right) \right]$,
    \item $\mathbf{0}$ is the $2 \times |H|$ zero-matrix, and
    \item $\mathbf{I}_2$ is the $2 \times 2$ identity matrix.
\end{itemize}

Now, we generalize the previous notation of probabilities as $P^{(n)}$ i.e. the probability at depth-$n$ of the Graph, with $P$ implicitly referring to $P^{(1)}$. Then, let $h_s$ and $h_t$ be given source and target transient states in the Graph respectively. We further define the probability of \textit{success} of $h_t$ given $h_s$ such that $h_t$ is \textit{reached} by $h_s$ in at most $k$ steps as follows:

\begin{equation}
    \Phi_k\left(h_t\right) = P^{(1)}\left( r^+ | h_t \right) \cdot \sum\limits_{n = 0}^k P^{(n)}\left( h_t | h_s \right)
\end{equation}

\noindent As such, in the context of reducing defects, we consider $h_t$ to be a possible reformulation candidate for $h_s$ if it is \textit{reachable} by $h_s$, such that conditioned on $h_s$, $h_t$ has a higher chance of \textit{success} than $h_s$ on its own, i.e.:

\begin{equation}
    \label{inequality1}
    \begin{split}
       P^{(1)}\left( r^+ | h_t \right) \cdot \sum\limits_{n = 0}^{\infty} P^{(n)}\left( h_t | h_s \right) &> P^{(1)}\left( r^+ | h_s \right) \\
       \Phi_{\infty}(h_t) &> \Phi_1(h_s)
    \end{split}
\end{equation}

Here, \textit{reachability} of any two states implies that there exists a path between them in the Graph or mathematically speaking, there exists a non-zero value of $n$ for which $P^{(n)}\left( h_t | h_s \right) > 0$. Now, consider the probability of \textit{success} of $h_t$ given $h_s$ such that $h_t$ is reached by $h_s$ in exactly $n$ steps. We would then have the following:

\begin{equation}
    P^{(1)}\left(r^+ | h_t \right) \cdot P^{(n)}\left(h_t | h_s \right) = P^{(1)}\left( r^+ | h_t \right) \cdot q^{(n)}_{s,t}
\end{equation}

\noindent where $q^{(n)}_{s,t}$ refers to the $(s,t)$-entry of the matrix $\mathbf{Q}^n$ ($\mathbf{Q}$ multiplied by itself $n$ times), which in turn refers to the probability of reaching $h_t$ from $h_s$ in exactly $n$ steps i.e. $ P^{(n)}\left( h_t | h_s \right)$. Expanding this to \textit{any} number of steps i.e. \textit{reachable} would thus allow us to reformulate the left set of terms in the inequality of Eq. \ref{inequality1} using matrix notations:
\begin{equation}
  \begin{split}
    \Phi_{\infty}(h_t) &= P^{(1)}\left( r^+ | h_t \right) \cdot \sum\limits_{n=0}^{\infty} P^{(n)}\left(h_t | h_s\right) \\
    &= P^{(1)}\left( r^+ | h_t \right) \cdot \sum\limits_{n=0}^{\infty} q^{(n)}_{s,t}  \\
    &= P^{(1)}\left( r^+ | h_t \right)\cdot \left(\sum\limits_{n=0}^{\infty} \mathbf{Q}^n \right)_{s,t} 
  \end{split}
\end{equation}

Generalizing this across all $h \in H$, define the matrix $\mathbf{P}$ such that its $(s, t)$-th entry, $p_{s,t} = \Phi_{\infty}(h_t)$. Then, we have:

\begin{equation}
    \mathbf{P} = \left( \sum\limits_{n=0}^{\infty} \mathbf{Q}^n \right) \mathbf{R}^+_{dg}
\end{equation}

\noindent where $\mathbf{R}^+_{dg}$ is the diagonal matrix whose diagonal is the vector $\mathbf{r}^+$. Now, as $\mathbf{Q}$ is a square matrix of probabilities, we have $\lVert\mathbf{Q}\rVert < 1$ and that $\mathbf{Q}$ is convergent. Then the summation above leads to a geometric series of matrices, which as given by Definition 11.3 in \cite{grinstead}, corresponds to the fundamental matrix of the Markov model, denoted by $\mathbf{N}$:

\begin{equation}
    \mathbf{N} = \sum\limits_{n=0}^{\infty} \mathbf{Q}^n = \left(\mathbf{I}_{|H|} - \mathbf{Q}\right)^{-1}
\end{equation}

\noindent with $\mathbf{I}_{|H|}$ referring to the identity matrix with the dimensions, $|H| \times |H|$. Given this, let $\mathbf{p}^{(s)}$ be the $s$-th row vector of the matrix $\mathbf{P}$ corresponding to $h_s$. As such, every non-zero entry $t$ in $\mathbf{p}^{(s)}$ translates to the probability $\Phi_{\infty}(h_t)$ of some \textit{reachable} $h_t$. This vector is thus given by:

\begin{equation}
    \mathbf{p}^{(s)} = \left( \mathbf{N} \mathbf{R}^+_{dg} \right)_s = \mathbf{N}_s^{\top} \circ \mathbf{r}^+
\end{equation}

\noindent where $\circ$ refers to the Hadamard (\textit{element-wise}) product. We then frame our objective as identifying the $h_t$ which maximizes the aforementioned probability for the given $h_s$:

\begin{equation}
h_t^* = \text{arg}\max\limits_{h_t} \Phi_{\infty}(h_t) = \text{arg}\max\limits_t \mathbf{p}_t^{(s)}
\end{equation}

Intuitive speaking, in the event that $h_t^* \neq h_s$, the model shows that there exists a \textit{reachable} target interpretation that when reformulated from $h_s$, has a better chance at a \textit{successful} experience than not doing so. In reference to Fig. \ref{fig:graphfig}, we can see that reformulating $h_0$ to $h_t^* = h_2$ increases the likelihood of success as: 

\begin{equation}
    P^{(1)}\left(r^+ | h_2 \right) \cdot \sum\limits_{n=0}^{\infty} P^{(n)} \left( h_2 | h_0 \right) = \frac{2}{3} > P^{(1)}\left( r^+ | h_0 \right) = 0
\end{equation}

Suppose that $h_t^* = h_s$. In which case, the source interpretation is already \textit{successful} on its own and hence requires no reformulation. As such, the model is effectively able to automatically partition the vertex space, $H$ into sets of \textit{successful} ($H^+$) and \textit{unsuccessful} ($H^-$) interpretations. In extending this reformulation back to the utterance space, $U$, we leverage the distributions $P(U|H)$ and $P(H|U)$ defined in Eq. 5 and re-define our objective as follows for a given source utterance $u_s \in U$:

\begin{equation}
    \label{sumProbabilities}
    \begin{split}
        u_t^* &= \text{arg}\max_{u_t} \sum\limits_{h_s} \sum\limits_{h_t} P^{(1)}\left(u_t | h_t \right) \cdot \Phi_{\infty}(h_t) \cdot P^{(1)}\left( h_s | u_s \right)
    \end{split}
\end{equation}

\noindent The intuition described above can similarly be applied here where $u_t^*$ is the more \textit{successful} reformulation of $u_s$. Note that the self-partitioning feature of the model directly extends to the utterance space, $U$, allowing it to surgically target only utterances that are likely to be defective and surface their corresponding rewrite candidates. This is the cardinal aspect of the model that drives the \textbf{self-learning} nature of the proposed system without requiring any human in the loop.

\subsection{Implementation}
With $|H| \sim 10^6$, constructing the matrix $\mathbf{Q}$, let alone inverting it, poses a key challenge towards scaling out the model, particularly in its batched form. As such, we formulate an approximation in computing the vector $\mathbf{p}^{(s)}$ for all source interpretations, $h_s$ by means of a distributed approach.

We note that from our dataset, $D^{\prime}$, that in the event that a given source utterance, $u_s$ is defective, users would only attempt at reformulating their query a few times before either arriving at a satisfactory experience or abandoning their session entirely. This translates to most ($\sim 97.3\%$) source interpretations, $h_s$ in the Markov model having short path lengths (\textit{i.e. typically $\leq 5$}) prior to them being absorbed by an absorbing state. Consequently, this along with the fact that these reformulations are recurrent across users, most high-confidence reformulations often only involve visiting a much smaller set of target interpretations, $h_t$, i.e.

\begin{equation}
    \sum\limits_{t=0}^{|H|} \mathbf{1}\left( \mathbf{p}_t^{(s)} > 0 \right) \ll |H|
\end{equation}

This leads us to deduce that the matrix $\mathbf{Q}$ is highly sparse and the corresponding Graph contains many clustered (\textit{i.e. community}) structures. We then leverage these facts to first collect the paths for every source interpretation, $h_s$ in a series of map-reduce tasks, by means of a distributed breadth-first search traversal up to a fixed depth of 5 using Apache Spark \cite{zaharia}. Thereafter, each task receives the paths corresponding to a single $h_s$ and in turn uses them to construct an approximate transition matrix, $\bar{\mathbf{A}}^{(s)}$. As the dimensionality of the matrix $\bar{\mathbf{A}}^{(s)}$ is much lower than that of $\mathbf{A}$, we can easily compute the approximate fundamental matrix, $\bar{\mathbf{N}}$ and the approximate vector $\bar{\mathbf{p}}^{(s)}$ within the same task. As a result, we have a distributed solution for parallelizing the computation of $\bar{\mathbf{p}}^{(s)}$ for every $h \in H$.

The breadth-first search traversal, which involves a series of sort-merge joins, does indeed introduce an algorithmic overhead of $\mathcal{O}(d \cdot |E| + |E| \log |E|)$, where $d$ and $E$ refer to the depth of the traversal and the set of all edges in the Graph respectively. We do also note that as this is a distributed join, the incurred network cost due to data shuffles are omitted here for simplicity. That being said, these overheads are offset by the advantage of being able to scale out the model. For purposes of optimization, each successive join is only performed on the set of paths which are non-cyclic and have yet to be absorbed while paths with vanishing probabilities are pruned off. 

\begin{table*}
    
    \caption{Some example rewrites from the Graph.}
    
    \label{table:exampleRewrites}
    \centering
    
    \begin{tabular}{cllc}
        \toprule
        \textbf{No.} & \textbf{Original utterance} & \textbf{Rewrite} & \textbf{Label} \\
        \midrule
        \textsf{1} & \textsf{play maj and dragons} & \textsf{play imagine dragons} & \multirow{8}{*}{\textbf{\textsf{Good}}} \\
        \textsf{2} & \textsf{play shadow by lady gaga} & \textsf{play shallow by lady gaga} &  \\
        \textsf{3} & \textsf{play rumer} & \textsf{play rumor by lee brice} &  \\
        \textsf{4} & \textsf{play sirius x. m. chill} & \textsf{play channel fifty three on sirius x. m.} &  \\
        \textsf{5} & \textsf{play a. b. c.} & \textsf{play the alphabet song} &  \\
        \textsf{6} & \textsf{don`t ever play that song again} & \textsf{thumbs down this song}  &  \\
        \textsf{7} & \textsf{turn the volume to half} & \textsf{volume five} &  \\
        \textsf{8} & \textsf{play island ninety point five} & \textsf{play island ninety eight point five} &  \\
        \midrule
        \textsf{9} & \textsf{play swaggy playlist} & \textsf{shuffle my songs} & \multirow{2}{*}{\textbf{\textsf{Bad}}} \\
        \textsf{10} & \textsf{play carter five by lil wayne} & \textsf{play carter four by lil wayne}  &  \\
        \bottomrule
    \end{tabular}
    
\end{table*}

\section{Experiments}

\subsection{Baseline: Pointer-Generator Sequence-to-Sequence Model}
Sequence-to-sequence (seq2seq) architectures have been the foundation for many neural machine translation and sequence learning tasks \cite{sutskever}. As such, by formulating the task of query rewriting as an extension of sequence learning, we used a Long Short-Term Memory-based (LSTM) model as an alternative method to produce rewrites. In short, we first mined 3 months of rephrase data using a rephrase detection ML model such that the first utterance was defective, and the rephrase was successful. We then used this data to train the seq2seq model, such that given the first utterance, it produces the second utterance. The model is based on well-established encoder-decoder architecture with attention and copy mechanisms \cite{see}. After the model is trained, we then used it to rewrite the same utterances that the Graph rewrites.

\subsection{Offline Analysis}
In order to evaluate the quality of the rewrites we obtained, we annotated 5,679 unique utterance-rewrite pairs generated using Graph, and estimated the accuracy and win/loss ratio to be 93.4\% and 12.0, respectively. Win/loss ratio is defined as the ratio of rewrites that result in better customer experience and the rewrites that deteriorate customer experience. We further used the seq2seq model to generate rewrite for these utterances as a baseline. 

Applying the seq2seq model on this dataset resulted in accuracy of 55.2\%, significantly lower than the accuracy of Graph. This is expected, since the Graph is 1) aggregating all three months of data (and not only rephrases), 2) taking into account the frequency of transitions whereas the seq2seq model only has unique rephrase pairs for training, and 3) utilizing the interpretation space to further compact and aggregate the utterances. However, the seq2seq model has the benefit of higher recall (\textit{since it can rewrite any utterance}), and it learns the patterns, e.g. \texttt{SongName} $\rightarrow$ play \texttt{SongName}. Another important difference between the Graph and seq2seq methods is that the Graph is capable of marking an utterance as \textit{Successful} i.e. when $h = h^*$. This is a signal to \textbf{not} rewrite an utterance, since on itself is mostly successful. However, the seq2seq model lacks this capability, and it may rewrite a successful utterance, and cause a friction.

Table \ref{table:exampleRewrites} shows some examples of good and bad rewrites from the Graph. It is clear from the examples that the rewrites are capable of fixing ASR (no. 1-3), NLU (no. 4-7) and even user errors (no. 8). On the other hand, there are cases that the rewrites fail (no. 9-10). One of the recurring cases of failure is when an utterance is rewritten to a generic utterance, like "play", or "shuffle my songs". This usually happens due to the original utterance not being successful, and the users trying many different paths that eventually \textit{loses information}, and is aggregated in a generic utterance (due to Eq. \ref{sumProbabilities}). Another common case of failure is when the rewrite changes the intention of the original utterance by changing the song name or artist name. This happens because of various reasons. For example, the data that we use for building the Graph may contain a period of time where the original utterance was not usually successful, so the users changed their mind by asking to play another similar song (like no. 10). The first type of error is easy to correct, by either applying rules or building a learning-based ranker after the Graph generation. The second type, however, is tricky to detect, since a lot of times, the change in the interpretation helps. We relied on an online blacklisting mechanism to remove these rewrites in the production system.

\section{Application Deployment}
\subsection{Offline Rewrite Mining}
Since there are thousands of new utterances per day, and there are constant changes to the upstream and downstream systems in Alexa on a daily basis, it is important to update our rewrites on a regular basis to remove stale and ineffective rewrites. We run daily jobs to mine the most recent rewrites in an offline fashion. This allows us to find the most recent rewrites and serve them to users. It is noteworthy that in case of conflicts between the rewrites, we pick the most recent rewrite, since it has the latest data. We have online alarms and metrics to monitor daily jobs, since sometimes changes to the upstream and downstream Alexa components can impact our rewrite mining algorithm. In case of large changes in our metrics, we do a dive deep into the data to find the root cause.

\subsection{Online Service}
Since the Graph is static during the period it is used, and there are many repetitive utterances per day, we opted to mine the rewrites as key-value pairs, where the original utterance is the key, and the rewrite is the value. For example, we store "play babe shark" $\rightarrow$ "play baby shark" as one entry. We then serve these pairs in a high-performance database to meet the low latency requirement. This allows us to decouple the offline mining process and the online serving process for high availability and low latency requirements.

\subsection{Online Performance}
After all the offline analysis and traffic simulations, we launched Graph rewrites in production in an A/B testing setup. We monitored the performance of our rewrites against no-rewrites for over two weeks, and we observed more than 30\% reduction in defect rate ($p\text{-value} < 0.001$), helping millions of users. We further measured the win/loss ratio three months after the release, by calculating the number of unique rewrites where rewriting is significantly better - win - or worse - loss - compared to no-rewrite option (we used \textit{Z}-test to test the significance, and set p-value threshold of 0.01). The post-launch win/loss ratio closely matched our offline estimate (11.8 online vs. 12.0 offline).

We have been running this application for over 9 months in production, and it has been serving millions of users since, improving their experience on a daily basis without getting in their way. We know this for a fact since we have been monitoring customer satisfaction metrics on a weekly basis. We monitor the total number of rewrites, and the average friction rate for the rewrites, along with average friction for no-rewrites. On top of tracking online metrics, we continue doing offline evaluations on a weekly basis, where we sample our traffic, and send it for annotation. Combining the online and offline metrics in a longitudinal fashion allows us to closely follow the changes in the customer experience, which is the ultimate metric for our system.

\section{Conclusion}

As conversational agents become more popular and grow into new scopes, it is critical for these systems to have self-learning mechanisms to fix the recurring issues continuously with minimal human intervention. In this paper, we presented a self-learning system that is able to efficiently target and rectify both systemic and customer errors at runtime by means of query reformulation. In particular, we proposed a highly-scalable collaborative-filtering mechanism based on an absorbing Markov chain to surface \textit{successful} utterance reformulations in conversational AI agents. Our system achieves a high precision performance thanks to aggregating large amounts of cross-user data in an offline fashion, without adversely impacting users' perceived latency by serving the rewrites in a look-up manner online. We have tested and deployed our system into production across millions of users, reducing customer frictions by more than 30\% and achieving a win/loss ratio of 11.8. Our solution has been customer-facing for over 9 months now, and it has helped millions of users to have a more seamless experience with Alexa.

\bibliography{aaai.bib}
\bibliographystyle{aaai}

\end{document}